# Illumination, Expression and Occlusion Invariant Pose-Adaptive Face Recognition System for Real-Time Applications

Shireesha Chintalapati[#1], M. V. Raghunadh[*2]

*Department of E and CE*
*NIT Warangal, Andhra Pradesh, INDIA-506004*

*Abstract*— Face recognition in real-time scenarios is mainly affected by illumination, expression and pose variations and also by occlusion. This paper presents the framework for pose-adaptive component-based face recognition system. The framework proposed deals with all the above mentioned issues. The steps involved in the presented framework are (i) facial landmark localisation, (ii) facial component extraction, (iii) pre-processing of facial image (iv) facial pose estimation (v) feature extraction using Local Binary Pattern Histograms of each component followed by (vi) fusion of pose adaptive classification of components. By employing pose adaptive classification, the recognition process is carried out on some part of database, based on estimated pose, instead of applying the recognition process on the whole database. Pre-processing techniques employed to overcome the problems due to illumination variation are also discussed in this paper. Component-based techniques provide better recognition rates when face images are occluded compared to the holistic methods. Our method is simple, feasible and provides better results when compared to other holistic methods.

## I. INTRODUCTION

In today's world the need for intelligent security systems is very much needed. Unauthorized usages of ID Number, PIN numbers or passwords are some of the security breaches existing in the world. This is because the above mentioned means does not define our true identity. Biometrics defines our true identity because they are difficult to forge. The identity of a person can be recognized from various biometrics such as fingerprints, face, iris which are physiological characteristics of a person which are difficult to be altered. A person's behavioural characteristics can also be considered as biometrics. Today's technology has evolved with various techniques to automatically identify and recognize the biometrics.

In this paper face is considered as the biometric. Face recognition is advantageous over other biometrics such as fingerprints and iris. Face region extraction is nonintrusive and is easily captured when compared to other biometrics. Face recognition is incorporated in applications related to surveillance, automated attendance management systems and many others. Rapid advancements have taken place in the field of face recognition owing to the advances in capturing devices, availability of huge database of face images in the internet and demand for automated applications and security.

Face recognition system mainly deals with Image Acquisition, Pre-processing, Face region detection and extraction, Feature extraction and Classification considering trained images. In the proposed system additional feature of pose estimation is incorporated. Face recognition can be classified based on the types of features such as Appearance-based features and Geometric-based features. Appearance-based features describe the texture of face where as Geometric based features describe the shape. The recognition rates of a face recognition system are mainly affected by illumination, expression and pose variations. The occluded faces also give poor recognition rates. In this paper Component-based face recognition system is discussed in which descriptors of each component extracted from face are fused together to recognize the probe image.

The organisation of this paper is as follows. Section II describes all the features in the proposed model in detail. In Section III the results of our experiment are discussed. Finally, in Section IV results are summarized and future work is proposed.

## II. LITERATURE REVIEW

This paper mainly deals with real time applications which are different from cooperative user scenarios. Computer login, e-passport are the examples of cooperative user scenarios in which user are in frontal pose and carry neutral expression on their face. But generally in real time scenarios the user's face is subjected to illumination, pose, and expression variations. Age, Occlusion and Distance from the capturing device may also affect the recognition rates. Face recognition can be classified into various categories depending on the approach to recognise the face. Geometric-based approach, Holistic-based approach, Local feature-based approach are some of the techniques employed. [1]

Geometric-based face recognition was proposed by R. Brunelli and T. Poggio [2]. The authors have proposed the use of geometric features such as width and length of nose, mouth position and chin shape are considered as face feature descriptors. Many other geometric-based approaches are proposed in literature, which use length, width and angle information of the facial points.

Higher dimensionality of face images is the major drawback in the recognition systems for Holistic-based approaches. Principal Component Analysis (PCA) reduces the dimension of the image by representing face image as the linear combination of the Eigen faces.[3] This type of representation maximizes the total variance in the training images. The face image which is represented in lower dimensional space is used as a feature descriptor. Classification of the probe image is carried out using Distance Classifier. Support Vector Machines are also used as classifiers for better classification. PCA approach does not take the discriminative information in the data into consideration which is the major drawback.

Linear Discriminative Analysis (LDA)[4] is another famous holistic approach which minimizes the within class variance while maximizing the variance between the different classes. This approach helps to consider the discriminative information in the trained data set. In real-time scenarios if the probe image is taken in different lighting conditions the recognition may go wrong with LDA because of lack of illumination information.

PCA and LDA would give better recognition rates in the cooperative user scenarios. In real-time scenarios instead of holistic approaches local feature based approaches are proposed. Instead of considering the whole high dimensional probe image only the local features of the image are used as feature descriptors. Local Binary patterns, Gabor Wavelets are popularly used for the feature extraction. [5]

Component-based approach has been proposed which is robust to occluded face recognition. To extract the components of the face different facial coordinates need to be identified. Active Shape Models (ASM), which makes use of shape of face, are used for face landmark localisation. Stasm open source software can be used for the implementation of ASM. Active Appearance Model (AAM) makes use of shape and texture of the face for landmark localisation. ASM and AAM require accurate placement of landmarks in annotation phase else it would result in wrong results. [6]

### III. DESCRIPTION OF PROPOSED SYSTEM

In this paper we propose a Component-based Pose adaptive Face recognition system. The various steps involved in the implementation of this system are (i) Face landmark localisation and Component Extraction, (ii) Pose Estimation of the facial region, (iii) Pre-processing, (iv) Descriptors of features extracted and their fusion, (v) Pose adaptive Classification of probe image based on trained images and lastly (vi) Database Development is also one of the important steps involved. The subsequent sections explain each step in detail.

*A. Face Landmark Localisation and Component Extraction*

In this experiment we have chosen 3D camera, Kinect sensor device, for capturing the scene. Kinect provides the 3D landmark data and it is inexpensive, portable and simple to use. Use of 3D imaging devices is advantageous over 2D imaging devices for landmarking. 2D landmarking is highly affected by pose variations and illumination effects. 3D landmarking techniques make use of curvature information and this helps to bridge the performance gap occurred in 2D case. Kinect helps to overcome the drawbacks of other 3D land marking approaches. ([7],[8]) Drawbacks due to other approaches are requirement of manual intervention to fix the facial landmarks and time requirements. Kinect makes use of 2D image and its Depth image acquired from Infra-red camera to find 2D landmarks and corresponding 3D points.

TABLE I
FACE FEATURE POINTS

| S. No. | Face Feature Point |
|---|---|
| 1. | Middle Top of Left Eyebrow |
| 2. | Under Mid Bottom of Left Eyelid |
| 3. | Right of left Eyebrow |
| 4. | Left of Left Eyebrow |
| 5. | Middle Top of Right Eyebrow |
| 6. | Under Mid Bottom of Right Eyelid |

| 7. | Right of Right Eyebrow |
|---|---|
| 8. | Left of Right Eyebrow |
| 9. | Tip of the Nose |
| 10. | Midpoint between Eyebrows |
| 11. | Left Corner of Mouth |
| 12. | Right Corner of the Mouth |
| 13. | Outside Left Corner of the Mouth |
| 14. | Outside Right Corner of the Mouth |
| 15. | Top Dip of the Upper Lip |
| 16. | Bottom of the Chin |
| 17. | Top Skull |
| 18. | Top Right of the Forehead |
| 19. | Middle point of the Forehead |
| 20. | Mid of the Right cheek |
| 21. | Mid of the Left Cheek |

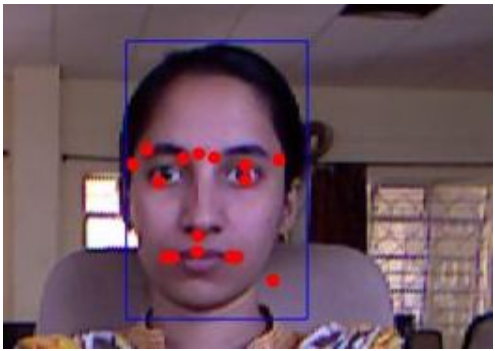

Fig. 1 Face Image with Feature Points

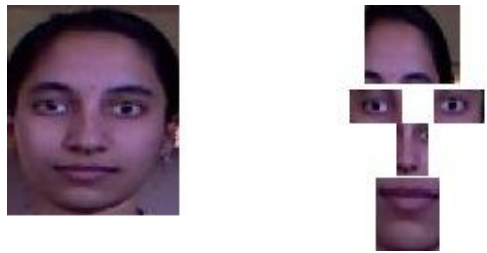

Fig. 2 Face Region and Components extracted

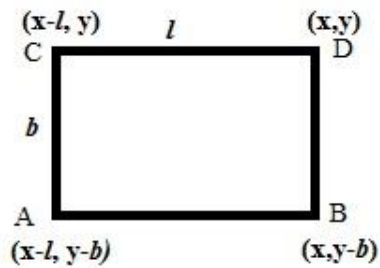

Fig. 3 Determination of Region of Interest from *l, b* and *P(x, y)*

In this experiment we have considered 21 feature points listed in Table 1 for component extraction and pose estimation. The components extracted from face region are forehead and eyebrow region, left eye, right eye, nose, mouth and chin region. These components are extracted from the facial region by setting a rectangular bounding box around the component as Region of Interest (ROI) and copying that portion of the image. The rectangular bounding box is determined from length (*l*), breadth (*b*) and right most pixel coordinates of the bounding box, say *P*(x, y). Therefore from *P*, *l* and *b* rest of the 3 vertices, which are the pixel coordinates on the image, are calculated and ROI is determined as shown in the Fig. 3. The length and breadth of the bounding box are calculated from the 2D pixel coordinates. For e.g. to find the length of mouth and chin region, the points Outside left corner of mouth, say P1(x1, y1) and Outside right corner of mouth, say P2(x2, y2) are considered and the distance between these two points gives the required measurement.

$$l = \sqrt{(x1 - x2)^2 + (y1 - y2)^2}$$

Similarly the breadth is measured using Right top dip of upper lip points, say P3 (x3, y3) and bottom of chin point, say P4 (x4, y4) using

$$b = \sqrt{(x3 - x4)^2 + (y3 - y4)^2}$$

In similar way the rest of the components of facial image are extracted from their *l*, *b* and *P* data. The facial image and the components of the corresponding image are as shown in the Fig. 2.

*B. Pose Estimation*

Pose Estimation plays a vital role in the face recognition techniques which are applied in real time scenarios. In this project the classification of the probe face image is accomplished using the database with images with similar head pose. This helps to reduce the time for classification. Pose dependent component classification is also employed. As we are using component-based face recognition techniques each component contribution is different in different poses. For e.g. when left profile face is facing the camera the contribution of right eyes is zero. Hence different databases are used for different poses. [13]

The head pose is basically divided into five categories namely Frontal, Left, Right, Up and Down. Each pose is determined depending upon Pitch, Yaw and Roll angles of the face image. The Kinect coordinate frame is as explained below. The Z axis is pointing towards the User, Y axis pointing Upwards and X axis to the right. [10] Pitch angle describes the upward and downward movement of face. Yaw angle describes the left and right

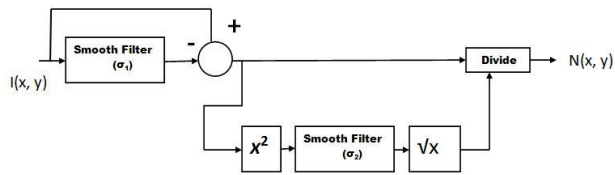

Fig. 4 Block Diagram for the implementation of Local Normalization

movement and Roll angle describes the movement of face with respect to the shoulder.

In order to determine the Pitch, Yaw and Roll angles the 3D depth face points obtained from Kinect camera are used. Pitch angle is angle obtained by subtracting 90 degrees from the angle between Z axis and the line joining Forehead centre and chin. Roll angle is the angle obtained by subtracting 90 degrees from the angle between X axis and the line joining Forehead centre and chin. Similarly Yaw angle is obtained from subtracting 90 degrees from the angle between Z axis and the line joining right and left cheeks.

The angle between a line joining two points, say $P1(x1, y1, z1)$ and $P2(x2, y2, z2)$, with respect to an axis, say Z axis is given by

$Angle = |z1 - z2|/\sqrt{((x1 - x2)^2 + (y1-y2)^2 + (z1-z2)^2)}$

If all the three angles are less than *25 degrees* and greater than *-25 degrees* then it is considered as frontal face. If Yaw angle is greater than *25 degrees* then it is right sided face and if it is less than *-25 degrees* then it is left sided face. Similarly if pitch is greater than *25 degrees* then it is up facing face and if less than *-25 degrees* then it is down facing face. 25 degrees is chosen as threshold from observation it can be changed for accurate results.

### C. Pre-processing of face image

Pre-processing steps are mainly employed to overcome the effect due to lighting variations on the face region. Local normalization is used to correct the non-uniform illumination on an image. It achieves it by uniformizing the mean and variance of an image around a local neighbour.

Consider an image $I(x, y)$ the normalized image $N(x, y)$ is given by

$$N(x, y) = (I(x, y) - \mu_I(x, y))/\sigma_I(x, y)$$

TABLE III
FIXED SIZE OF COMPONENTS

| Face Component | Size |
|---|---|
| Face | 92x112 |
| Left Eye | 27x18 |
| Right Eye | 27x18 |
| Nose | 24x38 |
| Mouth and Chin Region | 34x40 |
| Forehead and Eyebrow Region | 50x42 |

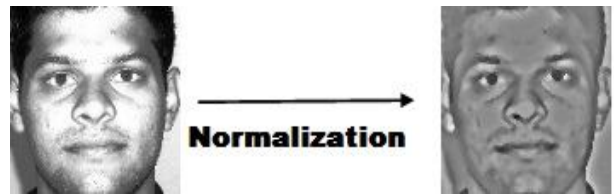

Fig. 5 Pre-processing for non-uniform illumination

where μ and σ are the mean and variance. The mean and variance are estimated using Gaussian spatial filters with different variances as shown in the Fig 4. The difference between the variances must not be small. [9] The figure Fig 5 shows the face image before and after normalization.

For training the images it is required to have all the images in same size. So the extracted ROIs are resized to a fixed size as shown in the Table 2.

### D. Feature Extraction

The feature descriptors of face and the components are described by the histograms of local binary patterns (LBP) of the respective images. The LBP face description implementation is computationally not complex and is robust to illumination variations. The LBP image is divided into local regions. The histograms of local regions are concatenated to represent the face or component descriptor.

LBP image of an input image is obtained in the following manner. Consider a 3x3 cell of an image as shown in the figure Fig 6. In order to obtain the local binary pattern the centre pixel intensity is compared with the neighbouring pixels and threshold is applied. If neighbouring pixel intensity is greater than the centre pixel intensity it is replaced by 0 else by 1. [11] The decimal number obtained from this binary pattern is used to represent the centre pixel. Mathematically LBP value at a pixel is given by the equations

$$LBP_c = \Sigma^{N-1}_{n=0} s(I_n - I_c) 2^n$$

where $I_n$ is the value at the neighbouring pixel and $I_c$ is the value of the centre pixel and $s(x)$ is given by

$$s(x) = 1, \text{if } x \geq 0;$$
$$= 0, \text{otherwise}$$

Likewise all the pixels are replaced using LBP codes which then gives rise to a LBP image. As we obtain LBP codes by comparing each pixel with neighbouring pixels we obtain a LBP image which is almost same for different illuminations as shown in Fig 7.

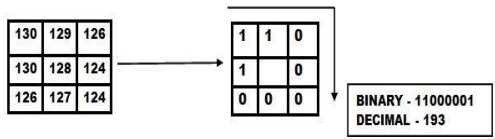

Fig. 6  LBP Pattern of a 3x3 mask

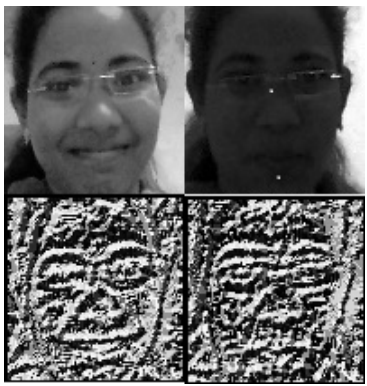

Fig. 7  LBP Image of gray images at different illumination

Now the obtained LBP image is divided into local regions and the histogram of each region is obtained. All these histograms are concatenated to get the face descriptor. Same technique is applied for all the components to obtain their respective descriptors.

*E.  Database Collection*

Database collection is another vital step in face recognition. For this experiment we have collected 8 different expressions of each person in 5 different poses, Frontal, Left, Right, Up and Down as shown in the Fig 8. 40 images of 120 people are collected for this project.

With respect to five different poses five subsets of database are collected. The components' contribution varies for each pose. So for Frontal database all the 5 components and face are collected. For Left database face and components excluding right eye are collected. For Right database face and components excluding left eye are collected. For Up database face and components excluding forehead portion are collected. Finally for Down database face and components excluding mouth region are collected. The next steps of processing are also implemented in similar manner.

So the database collected consists of faces with different expressions and poses. The pre-processing and feature extraction steps remove the hindrances caused by lighting variations.

*F.  Classification and Fusion*

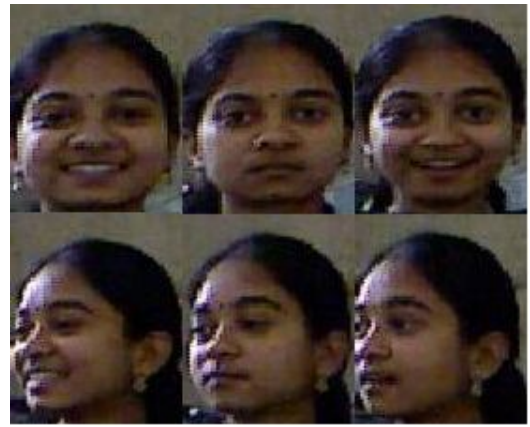

Fig. 8  Database with different poses and expressions

The face region of the input image is considered as probe image. The pose of the probe image is estimated and components are extracted depending up on the pose. The probe image is subjected to pre-processing.

The classification of the probe image using face and required number of components is accomplished employing sum of scores fusion. In this experiment features of different components have been extracted and these features are used for the classification. Similarity scores of each component is obtained. These scores need to be fused to get a final result. Different fusion techniques are present in the literature. We have used sum of scores fusion where each score has a weight assigned to it and the weighted sum results in the final score which gives the name of the recognized person.

$$s = \Sigma^N_{i=1} w_i s_i$$

where $w_i$ is the weight assigned to a individual score and $s_i$ is the score of a component.

Weights are assigned in such a manner depending upon the contribution of each component in different poses. For e.g. the contribution of right eye for a left profiled face is less, so it can be assigned a weight of 0. The weights assigned must follow the following criteria (i) weights belong to [0, 1] and (ii) sum of all weights is equal to 1. [12]

## IV. EXPERIMENTAL RESULTS

The face recognition system is developed in Microsoft Visual C# using EmguCV, C# wrapper of opencv. Kinect is used as a capturing device and it is placed at an average height of a person. Kinect device can only track 2 persons at a time. So the preferable number of persons in front of the camera is two. The colour image and depth image are captured at a resolution of 640x480 with 30frames per second.

### A. Component-based Evaluation

The recognition rate of different components, the result obtained from fused components and holistic face recognition results are presented in the Table 4. The database of 100 persons is collected for the evaluation of True Acceptance Rate (TAR). Only frontal faces are considered in this section.

TABLE IIIV
RECOGNITION RATE

| Feature | Recognition Rate |
|---|---|
| Left Eye | 60% |
| Right Eye | 61% |
| Nose | 65% |
| Mouth and Chin Region | 70% |
| Forehead and Eyebrow Region | 72% |
| Proposed Method | 92% |
| Holistic Approach | 85% |

### B. Under Pose Variations

In this section the performance of the face recognition system is evaluated under different poses. Table 5 presents the performance of holistic approach and fused component approach. The database of 100 persons with 40 images of each in different poses is considered.

TABLE V
RECOGNITION RATES AT DIFFERENT POSES

| Pose | Recognition Rate |
|---|---|
| Frontal | 92% |
| Left | 89% |
| Right | 89% |
| Up | 85% |
| Down | 80% |

### C. Under Expression Variations

The performance of under different expressions is evaluated. Expressions like smile, anger, talking are considered for the evaluation purpose. Results are shown in the Table 6.

TABLE VI
RECOGNITION RATES UNDER VARYING EXPRESSIONS

| Expressions | Holistic Method | Proposed Method |
|---|---|---|
| Smile | 89% | 91% |
| Anger | 89% | 91% |
| Talk | 85% | 90% |

### D. Under Occlusion

The faces to be recognized are covered partially to evaluate the performance of the system. The occluded faces are not collected in the database. Scarf to cover the lower portion of face and Sunglasses to cover the upper portion are used for the evaluation purpose.

TABLEV IVI
RECOGNITION RATES UNDER OCCLUSION

| Occlusion | Holistic Methods | Proposed Method |
|---|---|---|
| Sunglasses | 2% | 60% |
| Scarf | 1% | 55% |

## V. CONCLUSIONS AND FUTURE SCOPE

Face Recognition is being used in applications such as surveillance and attendance management system where most of the scenarios are non cooperative. In such cases the face recognition system has to be robust to be robust to illumination, pose and expression variations and also occlusions. The proposed method gives better recognition rates compared to holistic approaches. Component based feature extraction is employed in this project as it provides better results in case of occlusions. The contribution of components varies with respect to the pose, so pose adaptive classification is employed.

The recognition is good at short distances than compared to the long distances. In order to develop a robust human recognition system for surveillance application face features can be fused with the gait features for classification. Kinect can only track two persons at a time. Improvement of this feature helps in recognition of large number of people at a time.